RareAlert: Aligning heterogeneous large language model reasoning for early rare disease risk screening

Xi Chen[1], Hongru Zhou[2], Huahui Yi[3], Shiyu Feng[3,4], Hanyu Zhou[5], Tiancheng He[6], Mingke You[1], Li Wang[1], Qiankun Li[6], Kun Wang[6], Weili Fu[1], Kang Li[1], Jian Li[1]

[1]Sports Medicine Center, Department of Orthopedics and Orthopedic Research Institute, West China Hospital, Sichuan University, Chengdu, China
[2]Center for Cleft Lip and Palate Treatment, Plastic Surgery Hospital, Chinese Academy of Medical Sciences and Peking Union Medical College, Beijing, China.
[3]West China Biomedical Big Data Center, West China Hospital, Sichuan University, China
[4]The Chinese University of Hong Kong, Shenzhen, China
[5]School of Computer Science, Carnegie Mellon University, Pittsburgh, Pennsylvania, USA
[6]College of Computing and Data Science (CCDS), Nanyang Technological University, Singapore.

Jian Li: lijian_sportsmed@163.com

**Abstract**

Missed and delayed diagnosis remains a major challenge in rare disease care. At the initial clinical encounters, physicians assess rare disease risk using only limited information under high uncertainty. When high-risk patients are not recognised at this stage, targeted diagnostic testing is often not initiated, resulting in missed diagnosis. Existing primary care triage processes are structurally insufficient to reliably identify patients with rare diseases at initial clinical presentation and universal screening is needed to reduce diagnostic delay. Here we present RareAlert, an early screening system designed for universal application at all primary clinical encounters, which estimates patient-level rare disease risk from routinely available primary-visit information alone by leveraging heterogeneous medical reasoning from large language models (LLMs). RareAlert integrates reasoning generated by ten LLMs, calibrates and weights these signals using machine learning, and distils the aligned reasoning into a single locally deployable model. To develop and evaluate RareAlert, we curated RareBench, a real-world dataset of 158,666 cases covering 33 Orphanet disease categories and more than 7,000 rare conditions, including both rare and non-rare presentations. The results showed that rare disease identification can be reconceptualised as a universal uncertainty resolution process applied to the general patient population. On an independent test set, RareAlert, a Qwen3-4B based model trained with calibrated reasoning signals, achieved an AUC of 0.917, outperforming the best machine learning ensemble and all evaluated LLMs, including GPT-5, DeepSeek-R1, Claude-3.7-Sonnet, o3-mini, Gemini-2.5-Pro, and Qwen3-235B. These findings demonstrate the diversity in LLM medical reasoning and the effectiveness of aligning such reasoning in highly uncertain clinical tasks. By incorporating calibrated reasoning into a single model, RareAlert enables accurate, privacy-preserving, and scalable rare disease risk screening suitable for large-scale local deployment.

## Introduction

Rare diseases are defined by low prevalence, typically affecting between 5 and 76 individuals per

100,000[1], yet more than 7,000 distinct rare diseases have been described across 33 major disease categories[2, 3]. Despite their diversity, rare diseases share a common clinical challenge: delayed or incorrect diagnosis. More than 70% of patients with rare diseases experience at least one misdiagnosis, and for many conditions, patients undergo repeated diagnostic evaluations over many years before receiving a definitive diagnosis[4]. Prolonged diagnostic delays are associated with disease progression, irreversible organ damage, and missed opportunities for timely intervention.

A major contributor to diagnostic delay is the failure to recognize rare disease risk during the patient' s initial clinical encounter. In routine practice, decisions made at the primary visit determine whether further diagnostic investigations, such as genetic testing or specialist referral, are pursued. However, early manifestations of rare diseases are often nonspecific, overlap substantially with common conditions, and frequently involve symptoms across multiple organ systems. At this stage, clinicians must rely on limited information, typically restricted to demographic characteristics, medical history, and physical examination findings. Under these constraints, distinguishing patients who may harbor a rare disease from the much larger population with common disorders is inherently difficult, particularly in healthcare settings with limited resources or limited exposure to rare conditions[5-7].

Recent advances in artificial intelligence have shown promise in rare disease diagnosis. Prior studies, including ours, have demonstrated that large language models (LLMs) encode substantial medical knowledge, including information relevant to rare diseases, and can generate clinically coherent diagnostic reasoning[8, 9]. In parallel, machine learning models trained on structured clinical data have been used to predict specific rare diseases[10, 11].

However, important limitations persist in existing artificial intelligence–based approaches for rare diseases. Rare disease risk screening at the initial clinical encounter has received limited systematic investigation. Most studies relied on the results of further diagnostic investigations to establish rare disease diagnoses[11]. Practical considerations related to real-world deployment also remain underexplored, as many systems rely on computationally intensive or cloud-based models, raising concerns regarding cost, data privacy, and the feasibility of local deployment in routine clinical settings.

Rare disease patients cannot be reliably identified at the time of initial presentation using existing primary care triage processes. Universal screening is therefore impractical. Universal screening at the primary visit is the only viable strategy to reduce diagnostic delay. Therefore. we developed RareAlert, a universal screening system that can be applied to all patients presenting at primary clinical encounters. RareAlert operates exclusively on information routinely available during initial visits, including demographic data, free-text medical history, and physical examination findings, without reliance on laboratory tests, imaging studies, or specialized investigations. The system estimates patient-level risk across all major rare disease categories, encompassing more than 7,000 conditions, and provides quantitative risk assessments accompanied by interpretable clinical explanations to support clinician decision-making. RareAlert is designed for efficient local deployment, enabling large-scale screening while preserving data privacy and minimizing computational overhead.

A key challenge in early rare disease screening is the substantial variability in clinical reasoning when information is limited. Clinical reasoning often proceeds in a divergent manner, with multiple hypotheses considered in parallel before converging on a final conclusion. In complex cases, multidisciplinary collaboration (MDT) is commonly required to integrate complementary expertise[12].

Different clinicians may emphasize different features, generate distinct hypotheses, or assign varying levels of concern to the same presentation. Analogously, we observed that different large language models often follow heterogeneous medical reasoning paths when assessing rare disease risk for the same case. RareAlert leverages this diversity by integrating risk assessments and reasoning-derived signals from multiple LLMs with heterogeneous architectures and training data. These complementary signals are aligned and calibrated using machine learning–based methods, enabling the distillation of diverse reasoning perspectives into a single, deployable screening model.

Our study has several contributions to the field:

First, we introduce RareAlert, a universal screening system that can be applied to all patients presenting at primary clinical encounters. RareAlert estimates patient-level risk across all major rare disease categories, covering more than 7,000 conditions, and provides quantitative assessments to support decisions regarding further rare disease oriented diagnostic workup. The system is designed for efficient local deployment to enable large-scale screening while preserving data privacy.

Second, we propose a reasoning alignment framework that integrates diverse medical reasoning processes generated by large language models with heterogeneous architectures and training data. By treating model-generated reasoning paths as complementary signals and aligning them using machine learning based methods, this framework enables the distillation of multiple reasoning perspectives into a single deployable model. This approach provides a general strategy for incorporating heterogeneous reasoning into clinical decision support and is applicable across multiple rare disease related tasks.

Third, we construct RareBench, a large-scale, open-source real-world clinical dataset encompassing 158,666 cases for rare disease research. RareBench includes both rare disease and non-rare disease cases, spans all 33 major rare disease categories and more than 7,000 distinct conditions, and captures longitudinal clinical trajectories. Each case is structured according to real-world clinical workflows, including primary visit, follow-up visit, treatment planning, and prognosis assessment, providing a standardized foundation for developing and evaluating rare disease screening and management systems.

## Results

### RareBench

We constructed RareBench, a large-scale real-world clinical dataset comprising 158,666 curated cases, including 38,737 rare disease cases and 119,929 non-rare disease cases, to support the development and evaluation of rare disease diagnostic and management systems. The rare disease cohort spans all 33 Orphanet disease categories and covers more than 7,000 distinct rare conditions.

To assemble RareBench, we initially collected 432,811 clinical cases from multiple real-world sources. Rare disease candidates were obtained by mapping the complete Orphanet taxonomy to PubMed case reports, yielding 268,087 case report derived cases. Additional cases were sourced from MIMIC-IV-Ext-BHC, contributing 162,234 cases, and MIMIC-IV-Ext Clinical Decision Making, contributing 2,400 cases. All collected cases were processed through a standardized screening workflow, after which 158,666 cases met the inclusion criteria and were categorized into rare and non-rare disease cohorts.

For each included case, clinical information was organized according to the diagnostic and management pathway by constructing stage-specific representations with progressively increasing information density and distinct clinical tasks.

The Primary Visit represents the initial encounter at a primary healthcare facility and includes demographic information, medical history, and physical examination findings. The associated task is to assess rare disease risk, propose candidate diagnoses, and recommend further diagnostic evaluation.

The Follow-up Visit incorporates results from diagnostic investigations performed after the primary visit, including laboratory tests, imaging studies, and other examinations. The clinical task at this stage is to determine the final diagnosis.

The Treatment Visit adds confirmed diagnostic information and records clinical decision-making related to disease management, with the task of formulating an appropriate treatment plan.

The Prognosis Visit further includes information on implemented treatments and records patient outcomes, including long-term prognosis, functional status, symptom burden, and major clinical events. The task at this stage is to assess and predict patient prognosis.

**RareAlert**

RareAlert is an early screening system designed to estimate the risk of rare diseases at the point of primary clinical presentation. The system operates exclusively on information routinely available during initial encounters, including demographic data, free-text medical history, and physical examination findings. It is formulated as a binary screening task that distinguishes rare from non-rare disease cases across 33 disease categories, encompassing more than 7,000 rare diseases.

RareAlert integrates heterogeneous medical reasoning processes generated by multiple large language models (LLMs), aligns these reasoning-derived signals through machine learning based calibration, and distills both calibrated risk information and reasoning diversity into a single deployable model. The system outputs a quantitative estimate of rare disease risk, accompanied by a structured clinical explanation highlighting key contributing factors, and supports downstream classification into high-risk and low-risk groups using a predefined operating threshold determined during development. In the following sections, we describe the data foundation and system architecture underlying RareAlert, and then outline how multi-LLM medical reasoning and alignment are incorporated into the final model. The detailed development and evaluation pipeline is shown in methods section. An example of how RareALert aligned medical reasoning from multiple LLMs and provided the final risk estimate is shown in Figure 1.

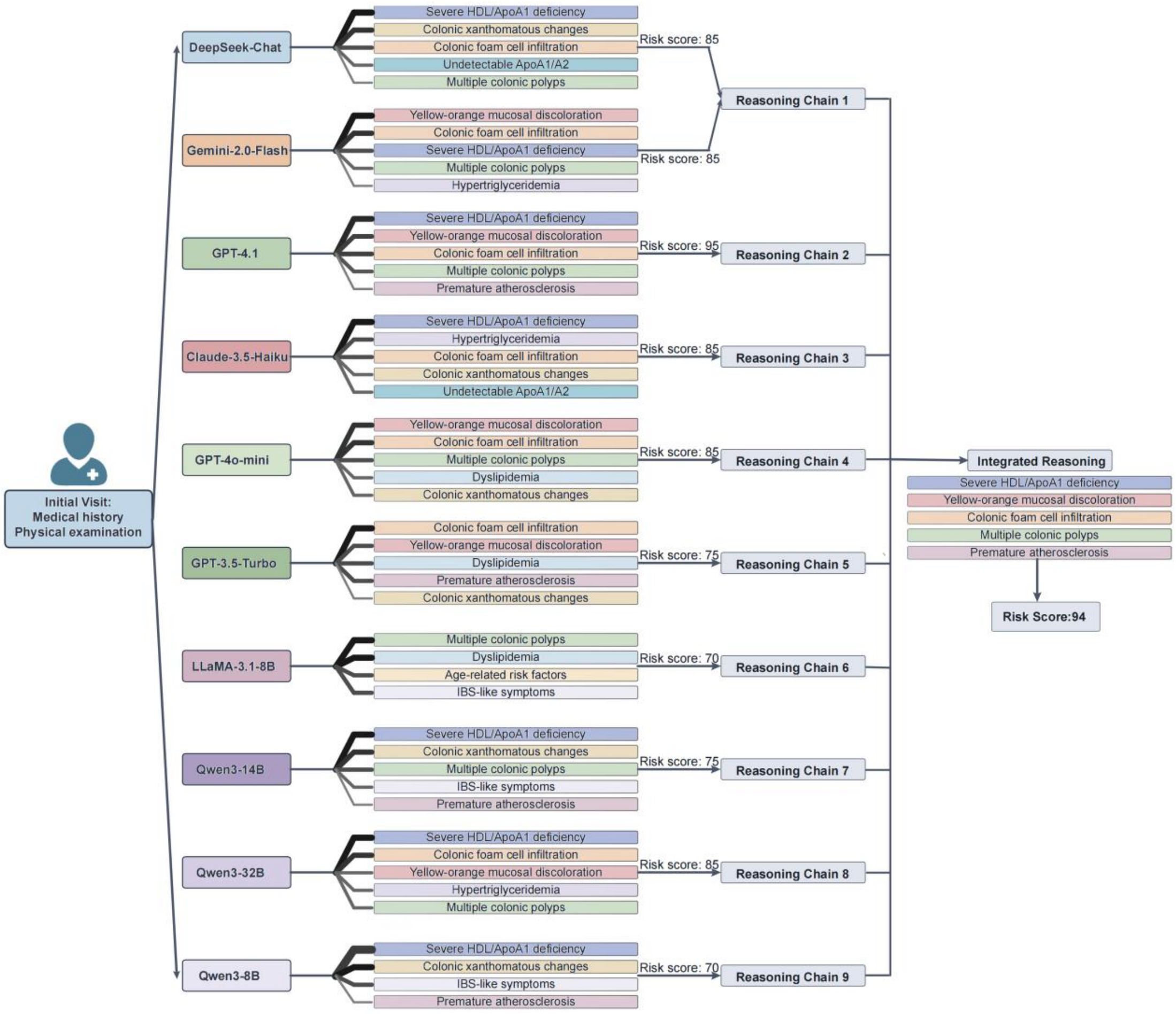

Figure 1. An example of how RareALert aligned medical reasoning from multiple LLMs and provided the final risk estimate.

**Stage 1 Data preparation and task formulation**

The system was developed and evaluated using the RareBench dataset, which consists of 158,666 real-world clinical cases, including both rare disease and non-rare disease presentations. Cases were randomly partitioned using stratified sampling into development and test sets with an 8:2 ratio, ensuring balanced representation of rare and non-rare cases in both splits. All model development, selection, and alignment procedures were confined to the development set, whereas the test set was reserved exclusively for downstream evaluation.

For each case, the input to RareAlert comprised three components available at the primary visit: patient demographics, free-text medical history, and physical examination findings. No laboratory results, imaging studies, or specialized diagnostic tests were used at any stage.

**Stage 2 LLM-based rare disease risk assessment**

Ten LLMs were tasked to assess the risk of rare disease for all cases from the Rarebench Dataset. Based on pilot experiments, the following models were included: Claude-3.5-Haiku, DeepSeek-V3, Gemini-2.0-Flash, GPT-3.5-Turbo, GPT-4.1, GPT-4o-mini, LLaMA 3.1 – 8B, Qwen3 – 8B, Qwen3 – 14B, and Qwen3 – 32B. These models differ in terms of training data, underlying architectures, parameter sizes, and open-source versus proprietary status, providing a broad foundation for RareAlert development. To capture heterogeneity in medical reasoning, multiple LLMs were applied independently to each case. Each model was instructed to perform clinical reasoning based on primary-visit information and to generate a structured output comprising three elements: a quantitative rare disease risk score on a standardized scale, a set of key clinical features contributing to the assessment with associated importance weights, and a synthesized clinical explanation that articulated the reasoning underlying the risk estimate.

**Stage 3. Alignment of reasoning-derived signals through machine learning**

To consolidate heterogeneous reasoning-derived risk assessments, we developed machine learning based alignment models that operated on the set of LLM outputs for each case. In this framework, each LLM's risk score was interpreted as a quantitative manifestation of a distinct underlying reasoning process. The alignment models integrated these signals to produce a calibrated reference risk estimate that reflected complementary information across reasoning pathways.

These alignment models were not intended for direct clinical deployment. Instead, they served as intermediate mechanisms for harmonizing multi-LLM reasoning and generating a stable supervisory signal for training the final RareAlert model. In addition to producing calibrated reference risk estimates, the alignment stage enabled case-specific attribution analyses that quantified the relative contribution of individual reasoning pathways to each prediction. This information was subsequently used to guide the selection of reasoning supervision during model distillation.

**Stage 4. RareAlert Training**

RareAlert was implemented as a single, locally deployable model that directly maps primary-visit clinical narratives to rare disease risk estimates with interpretable outputs. Model development was formulated as a supervised distillation process, in which the model was trained to reproduce the calibrated reference risk produced by the alignment stage while internalizing complementary medical reasoning patterns derived from multiple LLMs. Training was performed on the full development dataset, using a Qwen3-4B as the backbone and optimized via supervised fine-tuning.

During training, reasoning supervision was incorporated in a case-specific manner. For each case, medical reasoning outputs from multiple LLMs were provided as structured supervision signals, with selection informed by the relative importance of individual reasoning pathways estimated during alignment. This design allowed RareAlert to preferentially learn the most informative reasoning structures for each case, rather than uniformly mimicking all upstream models. Through this process, RareAlert internalized both a calibrated notion of rare disease risk and a repertoire of heterogeneous diagnostic reasoning strategies. At inference time, the model produces a quantitative risk estimate together with clinically interpretable explanations that highlight salient features driving the assessment, enabling transparent early screening at the point of initial clinical presentation.

**Performance Evaluation**

The performance of RareAlert was compared against individual large language models, machine learning based ensemble models developed during the alignment stage, simple statistical aggregation strategies (mean, median, and mode), and a multi-agent discussion framework. In addition, during model development, we systematically examined differences among LLMs in both quantitative risk scoring and underlying medical reasoning, including variability in predicted risk distributions, inter-model agreement and disagreement, and diversity of diagnostic reasoning paths across cases. We also examined how the optimal reasoning paths selected during RareAlert training affect model performance.

**Primary Outcome**

The primary outcome was the area under the receiver operating characteristic curve (ROC–AUC), which quantifies overall discriminative performance. On the independent test set, RareAlert achieved an AUC of 0.917, exceeding the best-performing machine learning ensemble (AUC 0.912) as well as all LLMs incorporated during model development, whose AUCs ranged from 0.680 to 0.896. RareAlert also outperformed advanced LLMs evaluated exclusively at the testing stage, including GPT-5, DeepSeek-R1, Claude-3.7-Sonnet, o3-mini, Gemini-2.5-Pro, and Qwen3-235B, which achieved AUCs between 0.880 and 0.897. In addition, RareAlert surpassed alternative ensemble techniques and multi-agent discussion approaches. Detailed comparative results are shown in Figure 1 and Table 1.

Table AUC Performance of RareAlert, Optimal ML algorithms, and LLMs

| Model | RareAlert | Optimal ML | GPT-5 | O3-mini | Deepseek-r1 | Claude-3.7-Sonnet | Gemini-2.5-Pro | Qwen3-235b | | |
|---|---|---|---|---|---|---|---|---|---|---|
| AUC | 0.917 | 0.912 | 0.905 | 0.897 | 0.88 | 0.896 | 0.882 | 0.859 | | |
| Model | GPT-3.5-turbo | GPT-4.1 | GPT-4o-mini | Llama3.1-8b | Deepseek-v3 | Claude-3.5-Haiku | Gemini-2.0-flash | Qwen3-32b | Qwen3-14b | Qwen3-8b |
| AUC | 0.691 | 0.896 | 0.679 | 0.746 | 0.808 | 0.798 | 0.804 | 0.809 | 0.829 | 0.749 |

**Secondary outcome measures**

Secondary outcome measures were evaluated at the decision threshold determined by the Youden index. Sensitivity, defined as the proportion of true rare disease cases correctly identified, was 0.778. This value was lower than that of the optimal machine learning ensemble and higher than those of all advanced LLMs evaluated at the testing stage, including GPT-5, DeepSeek-R1, Claude-3.7-Sonnet, o3-mini, Gemini-2.5-Pro, and Qwen3-235B. Specificity, defined as the proportion of non–rare disease cases correctly classified, was 0.935 and ranked highest among all evaluated models. Positive predictive value, which represents the proportion of predicted high-risk cases that were true rare disease cases, was 0.818 and ranked highest. Negative predictive value, reflecting the proportion of predicted low-risk cases that were true non–rare disease cases, was 0.918 and ranked second among all evaluated models.

We additionally assessed composite and correlation-based performance metrics. Overall accuracy, defined as the proportion of correctly classified cases, was 0.892. Balanced accuracy, defined as the mean of sensitivity and specificity, was 0.856. The F1 score, which balances precision and sensitivity, was 0.797, and the F2 score, which places greater weight on sensitivity, was 0.785. The Matthews correlation coefficient, a summary metric that incorporates all elements of the confusion matrix, was 0.668. For each of these metrics, RareAlert ranked highest among all evaluated methods. Detailed results are shown in Figure 2.

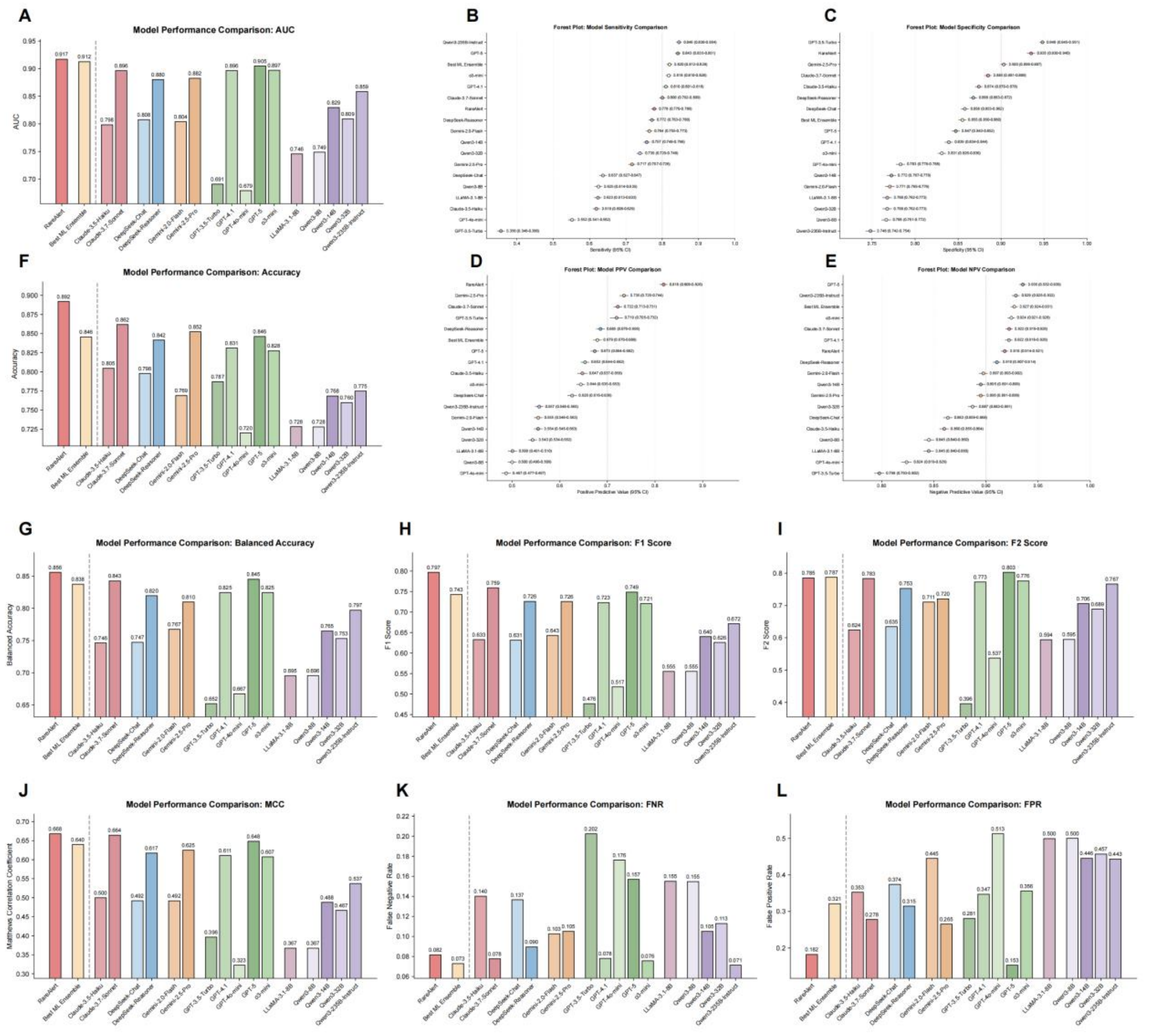

Figure 2 Performance comparison of RareAlert, individual large language models, and machine learning baselines for early rare disease risk screening.

a, Receiver operating characteristic area under the curve (ROC – AUC) for RareAlert, the best-performing machine learning ensemble, and individual large language models evaluated on the independent test set.

b, Sensitivity across all evaluated models at the selected operating threshold.

c, Specificity across all evaluated models at the selected operating threshold.

d, Positive predictive value (PPV) across all evaluated models at the selected operating threshold.

e, Negative predictive value (NPV) across all evaluated models at the selected operating threshold.

f, Overall classification accuracy across all evaluated models.

g, Balanced accuracy across all evaluated models.

h, F1 score across all evaluated models.

i, F2 score across all evaluated models.

j, Matthews correlation coefficient (MCC) across all evaluated models.

k, False negative rate (FNR) across all evaluated models at the selected operating threshold.
l, False positive rate (FPR) across all evaluated models at the selected operating threshold.

**Results from the Development Phase**

Score Variability of LLMs Adopted in Development Phase: We conducted a detailed analysis of the LLMs included in the development phase to examine differences in discriminative performance, risk score outputs, and diversity in medical reasoning. Across all models, the predicted rare disease risk scores were consistently higher for rare disease cases than for non-rare cases.

We further evaluated inter-model variability at the case level. Pairwise analyses showed that, for the same case, the probability that two models produced identical risk scores ranged from 5% to 49%, with most model pairs exhibiting agreement rates below 20%. Analysis of score differences revealed that the mean absolute difference in predicted risk scores between models ranged from 4 to 26 points.

At the case level, we quantified the distribution of model-assigned risk scores for rare and non-rare cases by counting the number of unique scores assigned by different models to the same case. For rare disease cases, four distinct risk scores per case were most frequently observed, accounting for 34.82% of cases, whereas for non-rare cases, five distinct scores were most common (33.42%). In terms of score magnitude, rare disease cases were predominantly assigned higher risk scores, with the largest proportions falling within the 85 – 90 (34.65%) and 75 – 80 (25.96%) ranges. In contrast, non-rare cases were more frequently assigned mid-range scores, most commonly within the 75 – 80 (26.04%) and 65 – 70 (25.24%) ranges. Detailed results are shown in Figure 3.

Reasoning Diversity of LLMs Adopted in Development Phase: To analyze diversity in medical reasoning across LLMs, we performed a post hoc analysis on a stratified random sample of 10,000 cases, comprising 2,500 rare disease cases and 7,500 non-rare cases. For each case, reasoning outputs from all ten LLMs were collected and analyzed using GPT-5 as an analysis-only model under deterministic settings. GPT-5 clustered the model outputs into distinct reasoning chains based on underlying diagnostic logic, summarized each chain, identified key differences in reasoning focus and causal interpretation, and quantified overlap across models.

The mean number of reasoning chains per case was comparable between rare and non-rare cases (6.017 ± 0.707 vs. 6.458 ± 1.414). The distribution of cases with different numbers of reasoning chains is shown in Figure 2, with six distinct reasoning chains per case being the most common scenario, accounting for 16.73% of cases. Further analysis demonstrated that LLMs tend to follow distinct medical reasoning pathways. Pairwise analyses indicated that the probability of two models co-occurring within the same reasoning chain was predominantly low, with most agreement rates below 25% and many below 15%, reflecting limited concordance across the majority of model pairs. The model consensus score was defined as the average number of models supporting the reasoning chains in which a given model participated. Detailed results are shown in Figure 3.

Performance of Different Machine Learning Algorithms and the Effect of Involving Different Combination of LLMs in Training: All candidate machine learning algorithms and feature combinations were explored and selected exclusively within the development cohort using cross-validation. After the modeling pipeline was finalized, the selected model was evaluated on an independent held-out test cohort that was not used at any stage of model development or model selection.

In the development set, CatBoost achieved the highest discriminative performance, with an area under the receiver operating characteristic curve (AUC) of 0.917. Other algorithms, including Explainable Boosting Machine (EBM), LightGBM, stacking ensembles, XGBoost, multilayer perceptron (MLP), AdaBoost, logistic regression, random forest, and ExtraTrees, demonstrated AUC values ranging from 0.863 to 0.917. As an increasing number of LLM-derived outputs were incorporated as input features, model performance showed a consistent improvement, with the AUC steadily increasing when expanding from one LLM feature to a combination of ten LLM features. Detailed results are shown in Figure 3.

SHAP-Based Attribution Analysis: Based on the final machine learning model, we conducted an interpretability analysis using SHapley Additive exPlanations (SHAP) to evaluate how different LLM-derived inputs contributed to rare disease risk prediction. The objectives of this analysis were to quantify the relative contribution of individual LLMs to risk prediction, to rank LLMs according to the importance of their medical reasoning, and to use the prioritized reasoning outputs to construct training data for the subsequent RareAlert model.

Normalized global importance was defined as the proportion of the total absolute SHAP attribution contributed by each feature across all samples. This metric reflects the relative contribution of each feature to overall model predictions and enables feature importance to be compared on a standardized scale. The results indicated that GPT-4.1 was the most influential contributor among all LLM-derived features. The global SHAP value was calculated as the mean signed SHAP value for each feature across samples, capturing the average directional effect of each LLM on the predicted risk. Positive values indicate a tendency to increase the predicted rare disease risk, whereas negative values indicate a tendency to decrease it. GPT-4.1 showed a consistent tendency to increase estimated rare disease risk, whereas the other models generally exhibited a tendency to decrease it. Finally, dominant-feature frequency was calculated by identifying, for each sample, the feature with the largest absolute SHAP value and computing the frequency with which each feature emerged as the primary driver of individual predictions. GPT-4.1 was identified as the dominant feature in 87.15% of cases, highlighting its central role in driving the final model's predictions. Detailed results are shown in Figure 3.

Performance of Fine-Tuning with Different Numbers of LLM-Derived Reasoning Chains: We evaluated the performance of RareAlert by varying the number of LLM-derived medical reasoning chains incorporated for risk prediction, ranging from a single dominant LLM to all ten LLMs. In the preceding machine learning stage, LLMs were ranked according to their SHAP importance for each case, and models were trained using the top n ranked reasoning chains, where n ranged from 1 to 10. When n equaled 1, only the most influential LLM was included, whereas for larger values of n, the outputs of the corresponding top-ranked LLMs were incorporated.

The results indicated that, when fine-tuning the Qwen3-4B model, using only the single most important reasoning chain yielded the best performance on the test set, achieving an AUC of 0.917. In contrast, models trained with two to ten reasoning chains demonstrated slightly lower and relatively stable performance, with AUC values ranging from 0.905 to 0.907. The detailed results are listed in the Table 2. Detailed results are shown in Figure 3.

Table 2 Effect of Different Numbers of Reasoning Paths Included in RareAlert Training

| **Top N Reasoning Paths** | N = 1 | N = 2 | N = 3 | N = 4 | N = 5 | N = 6 | N = 7 | N = 8 | N = 9 | N = 10 |
|---|---|---|---|---|---|---|---|---|---|---|
| **AUC** | 0.917 | 0.909 | 0.906 | 0.906 | 0.907 | 0.906 | 0.906 | 0.905 | 0.906 | 0.907 |

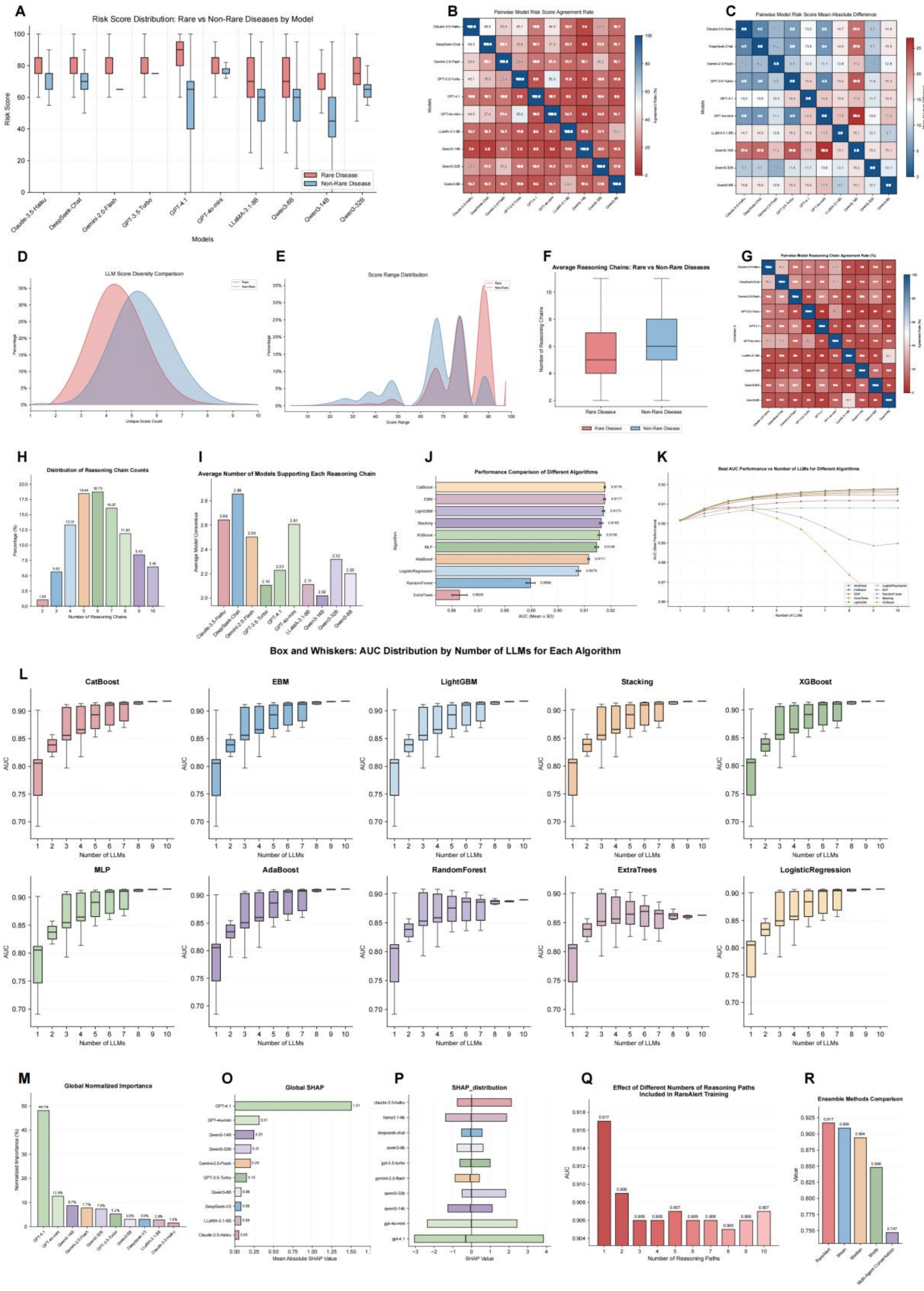

Figure 3 Results from developmental phase

A, Distribution of rare disease risk scores generated by individual large language models for rare disease and non-rare disease cases.

B, Pairwise agreement rates of risk scores between large language models, quantified as the percentage of cases in which two models assigned identical scores.

C, Pairwise mean absolute differences in risk scores between large language models across all cases.

D, Distribution of number of unique score generated per case for rare disease and non-rare disease cases.

E, Distribution of risk scores across predefined score ranges for rare disease and non-rare disease cases.

F, Number of distinct risk scores assigned by different models to the same case, shown separately for rare disease and non-rare disease cases.

G, Pairwise agreement rates of medical reasoning chains between large language models.

H, Distribution of the number of distinct reasoning chains identified per case.

I, Average number of models supporting each reasoning chain for individual large language models.

J, Comparison of performance (ROC–AUC) across different machine learning algorithms.

K, Best achieved ROC–AUC for different number of large language models included for different machine learning algorithms.

L, Effect of different number of large language models included in the training of each machine learning algorithms.

M, Global normalized importance for each large language model.

N, Average absolute SHPA value for each large language model .

O, Average SHPA value for each large language model.

P, Effect of incorporating different numbers of large language model–derived reasoning paths during RareAlert training, evaluated by ROC–AUC.

Q, Comparison of RareAlert with alternative ensemble strategies, including mean, median, mode aggregation, and a multi-agent conversation framework, evaluated by ROC–AUC.

Figure 3.

**Results of Other Ensemble Approaches and Multi-Agent Conversation**

We evaluated the performance of alternative ensemble strategies and a multi-agent conversation framework on the independent test set and compared their performance with that of RareAlert. The alternative ensemble methods included simple aggregation of risk scores generated by all LLMs using the mean, median, and mode. The multi-agent conversation approach was implemented using a previously established framework in which three physician agents, built on GPT-4o-mini as the base model, engaged in a structured discussion of each patient's clinical presentation and jointly produced a final rare disease risk score. The results demonstrated that RareAlert achieved the highest discriminative performance with an AUC of 0.917, outperforming mean-based ensemble aggregation with an AUC of 0.909, median-based aggregation with an AUC of 0.894, mode-based aggregation with an AUC of 0.848, and the multi-agent conversation approach with an AUC of 0.747.

## Discussion

Main Finding

Unlike previous studies, RareAlert is designed as a screening system rather than a diagnostic tool, with the explicit goal of supporting universal risk assessment at the initial clinical encounter. This design reflects a fundamental reality of rare disease care: patients with rare diseases cannot be reliably identified through upfront triage, as early presentations are often nonspecific and indistinguishable from common conditions. As a result, targeted screening limited to pre-suspected patients is impractical, and universal screening at the primary visit represents the only viable strategy for reducing missed diagnostic opportunities.

In this context, predictive performance should be interpreted in relation to downstream clinical actions, particularly the initiation and prioritisation of further diagnostic evaluation or specialist referral. The operating characteristics observed for RareAlert align with this intended role. High specificity and positive predictive value indicate that patients flagged as high risk are more likely to represent true rare disease cases, thereby helping clinicians focus diagnostic attention while limiting unnecessary investigations and controlling system-level false positives.

At the same time, the achieved sensitivity highlights the inherent trade-offs of early screening under information constraints. Some rare diseases present with subtle or ambiguous features that remain difficult to distinguish at the primary visit. In practice, the operating threshold of a screening system such as RareAlert can be adapted according to local resource availability, tolerance for false positives, and clinical priorities. Importantly, the model output is intended to support, rather than replace, clinician judgement, by providing an interpretable early risk signal that can raise diagnostic awareness and prompt more timely and targeted follow-up in appropriate cases.

Clinical interpretation and screening implications

Unlike previous studies, RareAlert is designed as a screening system rather than a diagnostic 

Alignment of Heterogeneous Reasoning Chains

The performance gains observed for RareAlert can be attributed to the complementary nature of medical reasoning generated by different large language models. Our results showed, when presented with the same clinical case, distinct LLMs often emphasise different symptoms, comorbidities, or disease categories, reflecting variations in training data, inductive biases, and reasoning strategies. Machine learning based alignment provides a principled mechanism to integrate this heterogeneous information. By calibrating and weighting reasoning signals at the case level, the alignment process selectively amplifies informative reasoning while attenuating noisy or misleading paths. Importantly, the observed improvements are not achieved by increasing the number of reasoning paths or the scale of inference at deployment. Instead, distilling aligned reasoning into a single compact model preserves the informative diversity present during training while avoiding the instability, cost, and latency associated with multi-agent inference. This finding underscores the importance of reasoning selection and alignment, rather than unfiltered aggregation, in high-uncertainty clinical prediction tasks.

Comparison with existing approaches

To the best of our knowledge, this is the first study targeted at rare disease identification at initial clinical encounters. Prior applications of artificial intelligence to rare disease have primarily focused on diagnosis, which relied on structured clinical data, often targeting a limited set of diseases or later stages of the diagnostic process when more comprehensive information is available[1-3]. While these approaches have demonstrated utility in specific contexts, they are less suited to the earliest clinical encounter, where information is sparse and the clinical task centres on risk stratification rather than diagnosis. While these approaches have demonstrated utility in specific contexts, they are less suited to the earliest clinical encounter, where information is sparse and the clinical task centres on risk stratification rather than diagnosis.

Recent advances in large language models have shown that single LLMs can perform complex medical reasoning on case-based clinical inputs[4, 5]. However, the performance of individual models can be variable, and their deployment as standalone screening tools in real-world workflows poses practical challenges related to reliability, cost, and integration.More recent work has explored strategies that aggregate multiple reasoning paths, including self-consistency and multi-agent debate, to improve robustness and reasoning quality[6, 7]. Our previous work showed that multi-agent debate can improve rare disease diagnostic accuracy. In contrast, applying multi-agent debate in this study did not improve rare disease risk assessment performance. We speculate that early-stage risk screening under limited information represents a distinct task that is underrepresented in LLM training, such that increasing inference-time deliberation or repeated sampling offers limited benefit. Consistently, in our early attempts, repeated risk estimation by a single LLM followed by aggregation (mean, median, or majority vote) also failed to improve screening performance.

RareBench Dataset

The development of RareAlert is enabled by RareBench, a dataset specifically constructed to support early-stage rare disease risk screening under limited information. RareBench comprises 158,666 real-world cases, spanning 33 Orphanet disease categories and more than 7,000 distinct rare conditions. To our knowledge, it represents the largest and most comprehensive collection of clinically grounded rare disease cases assembled for rare disease care, while explicitly including both rare and non-rare presentations. This composition allows systematic evaluation of false positives and false negatives, which is essential for screening-oriented tasks but is often not supported by existing rare disease datasets.

In addition to its scale, RareBench is organised around clinically meaningful stages, with primary visit information separated from downstream diagnostic, treatment, and outcome data. Although the present study focuses on the initial clinical encounter, this structure enables RareBench to serve as a foundation for developing and evaluating downstream systems, including diagnostic decision support, treatment planning, and longitudinal patient management. By aligning dataset design with the full clinical trajectory of rare disease care, RareBench supports not only the current risk screening task but also future research into end-to-end rare disease decision support under real-world constraints.

## Limitation

Several limitations of this study should be acknowledged. First, although RareBench comprises a large and diverse collection of clinically grounded cases, it is constructed from retrospective sources and predominantly reflects English-language clinical narratives. Differences in documentation practices, healthcare systems, and patient populations may affect model performance when applied in other settings, underscoring the need for external validation and site-specific calibration.

Second, the present work formulates rare disease identification as a binary risk screening task, without attempting to localise specific disease categories or directly recommend diagnostic tests. While this design aligns with the goal of early risk stratification at the primary visit, translating elevated risk signals into actionable clinical pathways remains an important challenge. To address this, we are developing downstream modules that operate on patients identified as high risk, with the aim of supporting targeted diagnostic reasoning, test prioritisation, and subsequent treatment decision-making. These efforts build on the staged structure of RareBench and are intended to extend RareAlert into an end-to-end rare disease decision-support framework.

Finally, our evaluation is limited to retrospective benchmarking and does not capture the real-world impact of RareAlert on clinical decision-making or patient outcomes. Prospective studies will be required to assess how rare disease risk alerts influence clinician behaviour, diagnostic timelines, and resource utilisation, as well as to evaluate fairness and performance across patient subgroups. Addressing these challenges will be essential for translating early risk screening models into routine clinical practice.

## Conclusion

In summary, this study demonstrates that early rare disease risk screening at the primary clinical visit is feasible using limited information alone. By aligning and distilling heterogeneous medical reasoning from multiple large language models, RareAlert achieves robust performance while remaining scalable and suitable for local, privacy-preserving deployment. Beyond its immediate clinical application, this work highlights the importance of reasoning diversity and task-specific alignment in high-uncertainty medical settings. Together with the RareBench dataset and ongoing development of downstream decision-support modules, RareAlert provides a foundation for end-to-end computational support across the rare disease care pathway.

## Methods

### Rare Benchmark Development

Previous research has established several benchmarks for rare diseases, such as xxxx. These benchmarks primarily assess diagnostic capabilities by testing whether language models can correctly diagnose rare diseases when complete patient information is provided. Despite their utility, existing benchmarks fail to meet our research requirements in three critical aspects:

First, our dataset requires the inclusion of negative cases (non-rare disease patients) to evaluate false positive rates in rare disease detection accurately. Second, to support our comprehensive rare disease management system, the dataset must provide broad coverage across all disease categories with sufficient case numbers for robust evaluation. Third, our benchmark necessarily incorporates treatment and prognostic information, as these elements are vital components for adequate clinical decision support.

To address these needs, we have developed a dataset comprising 158,666 real-world clinical cases, including 38,737 cases of rare diseases spanning all 33 rare disease categories and more than 7,000 distinct rare conditions.

### Data Acquisition

Data acquired from published case reports

Rare disease case reports were collected using the Orphanet database, a resource co funded by the European Commission that catalogues more than 7,000 rare diseases across 33 categories. After obtaining the complete disease list and corresponding classifications, we conducted systematic searches for relevant case reports in the PubMed database. Searches followed a standardized query format: "("disease"[MeSH Terms] OR "disease"[All Fields]) AND ("case reports"[Publication Type] OR "case reports"[All Fields])". For instance, when searching for Down syndrome cases, we used: "("down syndrome"[MeSH Terms] OR ("down"[All Fields] AND "syndrome"[All Fields]) OR "down syndrome"[All Fields] OR "trisomy 21"[All Fields]) AND ("case reports"[Publication Type] OR "case reports"[All Fields])". When a search returned more than 100 results, only the top 100 articles ranked by relevance were retrieved to maintain manageable data volume while preserving relevance. This process yielded 268,087 cases from published case reports.

Data Acquired from MIMIC Database

Additional cases were obtained from two MIMIC IV Ext resources. These included 162,234 cases from MIMIC IV Ext BHC, which contains longitudinal multi specialty records from a single medical center[13-15] and 2,400 cases from the MIMIC IV Ext Clinical Decision Making dataset, which focuses on emergency department visits[15-17].

Across all sources, 432,811 cases were collected for downstream screening.

Data Screen

All collected cases underwent a three stage screening workflow. First, duplicate cases were removed using case identifiers. Second, publications were assessed for eligibility as English language case reports when applicable. Third, each case was evaluated against five inclusion criteria: (1) presence of essential clinical documentation (history, physical examination, diagnostic evaluation, diagnosis), (2) adequacy of clinical detail within each section to support diagnostic reasoning, (3) clinical plausibility with respect to established medical knowledge and guidelines, (4) diagnostic evidence supporting the reported diagnosis, and (5) documentation of treatment and outcomes.

Screening was implemented using DeepSeek-V3 via API calls. To develop the automated screening pipeline, 500 cases were randomly sampled from the candidate pool and used for iterative prompt development and parameter tuning. Through repeated testing, we finalized the screening prompts, selected DeepSeek-V3 as the screening model, and fixed the temperature parameter at 0. Following screening, 158,666 cases met all inclusion criteria.

### Identification of Rare Disease

Rare disease status was determined for all screened cases using a composite identification procedure.

First, the final diagnosis for each case was extracted using DeepSeek-V3 via API calls. The full case report text was provided as input, and the model was instructed to extract the final diagnosis and output it in a standardized JSON format.

Second, extracted diagnoses were matched against rare disease entries from the Orphanet database using two complementary approaches: 1) Diagnosis matching was performed using exact string correspondence between case diagnoses and Orphanet disease names. 2) A retrieval augmented generation based semantic matching approach was implemented. All Orphanet disease entries were embedded using the text-embedding-large-3 model, and a FAISS vector index was constructed for efficient nearest neighbor retrieval. Case level disease names were embedded in the same vector space and used as query vectors. For each case, the top candidate Orphanet diseases and their similarity scores were retrieved and organized into a structured textual context. This context, together with the original case diagnosis, was provided to the GPT 5 nano model, which determined whether the case diagnosis matched any retrieved Orphanet disease. Cases judged as matched were labeled as rare diseases, and unmatched cases were labeled as non rare diseases.

Third, cases classified as non rare diseases after the first two steps underwent additional review to mitigate false negatives. This step addressed two scenarios: diagnoses corresponding to rare diseases with low textual similarity or incomplete naming, and diagnoses of rare diseases not included in the Orphanet database. We applied a hybrid approach combining GPT-5 and human evaluation. A random sample of 1,000 cases was independently assessed by GPT-5 and by human reviewers. Prompt design and model parameters were iteratively refined based on agreement between model and human judgments, and the finalized configuration was applied to the remaining cases.

Through this procedure, 38,737 cases were identified as rare diseases and 119,929 cases were classified as non rare diseases.

### Data Curation

All included cases were curated using an automated pipeline, with rare disease cases classified under Orphanet's disease classification system and by medical specialty. Each rare disease case was structured to represent a longitudinal clinical trajectory consisting of four stages with increasing information content and distinct clinical decision making tasks. Non rare disease cases were structured at the primary visit stage only, consistent with their role as negative controls for rare disease risk screening. The four clinical stages were defined as follows:

**Primary Visit.** This stage represents the initial patient encounter at a primary healthcare facility and includes demographic information, medical history, and physical examination findings. At this stage, the clinical objective is to assess whether the patient is at potential risk for a rare disease, to identify plausible candidate diagnoses, and to recommend appropriate further diagnostic evaluations to clarify the etiology.

**Follow-up Visit.** This stage extends the Primary visit by incorporating results from additional diagnostic evaluations, including laboratory tests, radiographic studies, and other relevant investigations. The associated clinical task is to interpret accumulated diagnostic evidence to establish or refine the final diagnosis.

**Treatment Visit.** This stage extends the Follow-up visit by adding confirmed diagnostic information. It represents the post-diagnostic phase in which the clinical objective is to formulate and initiate an appropriate treatment strategy tailored to the identified disease.

**Prognosis Visit.** This stage extends the Treatment visit by incorporating information on implemented therapeutic interventions. It represents the post-treatment phase, in which the clinical objective is to assess and predict patient prognosis, including anticipated disease course, potential complications, and long-term outcomes.

Data curation followed an iterative development process similar to that used for screening. A random sample of 500 cases was used to evaluate different language models, temperature settings, prompt designs, and the number of API calls per case. Based on these experiments, DeepSeek-V3 was selected with temperature set to 0. Each case was curated using six sequential API calls, and outputs were constrained to a stable predefined JSON schema.

**RareAlert Development**

RareAlert is designed as a screening system for rare disease. It addresses a fundamental clinical challenge in rare disease diagnosis: the inability to reliably recognize patients at high risk for rare diseases during their primary clinical encounters, when only limited information is available. In routine practice, the initiation of confirmatory diagnostic testing for rare diseases requires clinicians to first suspect rare disease risk based primarily on patients’ medical history and physical examination findings. However, identifying such risk under information-constrained conditions is inherently difficult, contributing to the high rate of delayed or missed diagnoses observed across rare diseases.

To meet this unmet clinical need, RareAlert was designed as an early screening and alert module that operates exclusively on information available at primary visits, including demographic data, free-text medical history, and physical examination findings, without reliance on laboratory results, imaging studies, or specialized tests. The module aims to quantitatively estimate a patient’s risk of having any rare disease, identify key contributing clinical factors, and provide interpretable explanations to support clinician decision-making.

RareAlert was developed to satisfy four core requirements. First, it must effectively identify high-risk patients using limited clinical information, enabling early recognition before confirmatory testing is considered. Second, it must support comprehensive risk assessment across the full spectrum of rare diseases, rather than focusing on a narrow subset, to ensure clinical utility across specialties and to reduce missed diagnoses in patients with multisystem involvement. Third, it must provide calibrated quantitative risk scores accompanied by clinically meaningful explanations, allowing physicians to interpret risk levels and underlying drivers in a transparent manner. Fourth, the system must be suitable for large-scale global deployment, including in resource-limited settings, which necessitates efficient inference and local, privacy-preserving deployment.

These design requirements informed several key methodological choices. First, because primary-visit data are predominantly unstructured free-text narratives, large language models (LLMs) were selected as the primary analytical framework for clinical feature extraction and reasoning. Second, pilot analyses revealed that different LLMs often follow distinct medical reasoning paths when evaluating the same case, and that their performance varies across patients. This observation suggested that integrating diverse LLM-derived reasoning signals could enhance rare disease risk prediction beyond any single model. Third, to enable scalable and globally deployable screening, the final system must rely on locally deployable open-source models.

Building on these considerations, RareAlert was developed through a multi-stage pipeline that combines multi-LLM risk assessment, machine learning – based score integration, and supervised distillation into a single deployable model for rare disease risk screening at the point of initial clinical presentation.

**Development Pipeline**

We developed RareAlert to quantitatively assess the risk of rare diseases using limited information available from patients' primary clinical visits. The module provides calibrated risk scores and clinically interpretable explanations across 33 disease categories, covering more than 7,000 rare diseases.

The overall development of RareAlert followed a multi-stage pipeline. In the first stage, clinical cases were collected and partitioned into development and test sets. In the second stage, multiple large language models (LLMs) were applied to each case to generate rare disease risk assessments based solely on primary-visit information. In the third stage, intermediate machine learning – based ensemble models were constructed to integrate heterogeneous risk signals derived from multiple LLMs, generate a calibrated reference risk score for each case, and estimate the case-specific importance of individual LLM outputs. These ensemble models were not intended as final predictive systems, but rather served as intermediate alignment mechanisms that consolidated complementary reasoning signals across models. In the fourth stage, RareAlert was trained using patient primary-visit information together with LLM-generated reasoning paths and ensemble-derived reference risk scores. The final model was trained to directly map free-text clinical narratives from primary visits to rare disease risk, while internalizing both the integrated risk signal and the diversity of LLM-derived medical reasoning.

After training, RareAlert accepts primary-visit clinical information as input and outputs a quantitative rare disease risk score, highlights key contributing clinical factors, provides an interpretable clinical explanation, and classifies patients as high risk or low risk according to the predefined threshold. In the final stage, the performance of RareAlert was systematically compared with individual LLMs, alternative machine learning–based ensemble approaches, and a multi-agent discussion framework.

**Stage 1: Data acquisition and partitioning**

This study utilized the Rarebench dataset, which comprises 158,666 real-world clinical cases, including 38,737 rare disease cases and 119,929 non-rare disease cases. Using stratified random sampling, cases from both groups were randomly allocated in an 8:2 ratio to form the development set and the test set, respectively. For the development of RareAlert, the information used for each case included three components: (i) primary-visit clinical information, consisting of patient demographics, medical history, and physical examination findings; (ii) a binary label indicating whether the case was a rare disease; and (iii) the specific rare disease category and associated medical specialty for rare disease cases.

**Stage 2: LLM-based rare disease risk assessment**

Selection of large language models

To capture heterogeneity in model architectures, parameter scales, and training paradigms, we selected a diverse set of large language models (LLMs) for rare disease risk assessment. Based on pilot experiments, the following models were included: Claude-3.5-Haiku, DeepSeek-V3, Gemini-2.0-Flash, GPT-3.5-Turbo, GPT-4.1, GPT-4o-mini, LLaMA 3.1–8B, Qwen3–8B, Qwen3–14B, and Qwen3–32B. These models differ substantially in terms of training data, underlying architectures, parameter sizes, and open-source versus proprietary status, providing a broad foundation for RareAlert development.

LLM evaluation setup

Each LLM independently processed all cases in the dataset, with a single API call corresponding to one case. For each evaluation, primary-visit clinical information was provided to the model via the context window. A standardized prompt instructed the LLM to perform rare disease risk assessment and to generate structured outputs consisting of three components.

First, the model was required to output a quantitative rare disease risk score as an integer ranging from 0 to 100, where 0 indicated no rare disease risk and 100 indicated certainty of a rare disease. Second, the model was asked to identify five key patient features that contributed to the assigned risk score. These features could include symptoms (for example, persistent fever), physical examination findings (for example, increased interocular distance), or elements of medical history (for example, a family history of hereditary disease). Each feature was assigned a weight reflecting its relative importance, with the weights summing to 1, and the most influential feature corresponding to the highest weight. The model was further instructed to provide an explanation for both feature selection and weight assignment. Third, the model was required to generate an overall clinical explanation synthesizing the risk score and feature-weight analysis.

To enhance output stability, the temperature parameter was set to 0 for all models. All outputs were constrained to a standardized JSON format to facilitate downstream processing.

Prompt optimization

LLM performance in rare disease risk prediction was optimized through prompt engineering. A pilot study was conducted using 750 randomly selected cases, including 250 rare disease cases and 500 non-rare disease cases. Two prompting strategies were evaluated. The first strategy used a direct input – output format, requiring the LLM to output only a rare disease risk score without explanation. The second strategy employed chain-of-thought reasoning, requiring the model to output the risk score along with key clinical features and their associated weights.

Across all evaluated LLMs, predictive performance was comparable among the three prompting strategies. We ultimately selected the second strategy, as it provided richer reasoning information that facilitated analysis of inter-model performance differences and enhanced clinical interpretability of model outputs.

Evaluation of LLM performance

We first evaluated the performance of each LLM in identifying patients at high risk for rare diseases. Model performance was assessed using evaluation metrics commonly employed in medical diagnostic studies, which are designed to determine whether a given score or indicator can effectively discriminate between diseased and non-diseased individuals. This evaluation framework is well aligned with our use of quantitative risk scores to assess rare disease risk.

The primary outcome metric was the area under the receiver operating characteristic curve (ROC – AUC), which measures a model's overall ability to discriminate between rare disease and non-rare disease cases across all possible decision thresholds. To define an operating threshold for classification, we applied the Youden index, a widely used criterion in diagnostic testing that identifies the threshold maximizing the sum of sensitivity and specificity. This threshold represents the optimal trade-off between true positive and true negative rates.

Based on the optimal threshold determined by the Youden index, we further evaluated model performance using sensitivity, specificity, false negative rate (FNR), false positive rate (FPR), positive predictive value (PPV), and negative predictive value (NPV). Sensitivity, equivalent to recall in machine learning terminology, reflects the proportion of true rare disease cases correctly identified. Specificity measures the proportion of non-rare disease cases correctly classified. The false negative and false positive rates quantify the proportions of missed rare disease cases and incorrectly flagged non-rare disease cases, respectively. PPV and NPV indicate the probabilities that patients classified as high risk or low risk truly have or do not have a rare disease. We additionally examined how sensitivity, specificity, PPV, and NPV varied as a function of the decision threshold to characterize model behavior across different screening scenarios.

We then assessed inter-model differences among LLMs, focusing on both variation in risk score outputs and differences in medical reasoning. With respect to risk score outputs, three analyses were conducted. First, we compared the mean predicted risk scores generated by each model for rare disease cases and non-rare disease cases, respectively. Second, we quantified inter-model score concordance by computing the probability that two models assigned the same risk score to the same case. Third, we measured inter-model risk score disagreement by calculating the pairwise mean absolute difference in risk scores assigned to the same case by different models.

Analysis of medical reasoning diversity across LLMs

To examine differences in medical reasoning across models, we conducted a post hoc analysis of LLM-generated reasoning outputs. To balance analytical depth with computational cost, a subset of cases was randomly selected for this analysis. Specifically, we applied stratified random sampling with a fixed 3:1 ratio of non-rare to rare disease cases, yielding a total of 10,000 cases, including 7,500 non-rare cases and 2,500 rare disease cases. Sampling was performed without replacement.

For each selected case, we collected the outputs generated by all ten LLMs evaluated in this study. We then used GPT-5 as an analysis-only model to perform post hoc examination of the reasoning structures produced by the other LLMs and to determine whether distinct medical reasoning paths were present for a given case.

Development of the reasoning analysis framework

The reasoning analysis framework was developed iteratively using a subset of 1,000 cases. During this phase, we selected GPT-5 as the analysis model and fixed the temperature parameter at 0 to ensure deterministic outputs. Prompt design was refined through multiple iterations, informed by comparisons between model-generated analyses and judgments provided by clinical experts, until stable and clinically coherent results were achieved.

For each case, GPT-5 received the complete set of reasoning outputs generated by all LLMs and was instructed to perform the following tasks. First, it identified distinct reasoning chains by clustering diagnostic reasoning outputs based on their underlying reasoning structure rather than surface-level wording. Reasoning chains were considered distinct if they differed in diagnostic focus, causal logic, feature prioritization, diagnostic style, or the scope and certainty of conclusions. Second, the model provided explicit justifications for differences between reasoning chains, describing how each chain diverged from the others in terms of primary diagnostic hypotheses, causal interpretations, or emphasis on specific organ systems. Third, each distinct reasoning chain was summarized to capture its core pathophysiological logic and interpretation of key clinical features, with an emphasis on reasoning structure rather than final diagnoses. Fourth, the model identified duplicated reasoning across LLMs by counting how many models produced each reasoning chain and recording the corresponding model identities. All outputs were returned in a standardized JSON format.

Quantification of reasoning diversity

Based on the identified unique reasoning chains for each case, we quantified the extent to which different LLMs followed distinct reasoning paths. The resulting metrics were conceptually grouped into three categories: core diversity metrics, structural similarity analyses, and auxiliary co-occurrence statistics.

Core diversity metrics included the number of reasoning chains per case, which reflects the diversity of reasoning outcomes within a case, and the number of models supporting each reasoning chain, which serves as a proxy for inter-model consensus. The probability of co-occurrence (P_same), defined as the likelihood that two models appeared together in a reasoning chain. Model consensus score for each LLM, defined as the average number of models supporting the reasoning chains in which that model participated.

**Stage 3: Alignment of reasoning-derived signals through machine learning**

In this study, we developed a machine learning framework to integrate risk assessments generated by multiple large language models (LLMs) at the case level. For each case, individual LLMs produced a quantitative score reflecting the likelihood of a rare disease. These LLM-derived risk scores were used as input features, while the ground-truth disease status (rare vs. non-rare) served as the binary outcome variable. All features and labels were converted to numeric format, and cases with missing outcome labels were excluded. To ensure comparability across different feature combinations and learning algorithms, model training and evaluation were conducted using complete-case analysis for the selected features, without imputation.

The trained ensemble models served as intermediate integrators of multi-LLM risk signals and provided calibrated reference risk scores that were subsequently used to supervise the training of the RareAlert model. Text-level ensemble methods were not adopted due to the difficulty of standardizing heterogeneous reasoning outputs across models. Instead, the informational value of LLM-derived reasoning paths was incorporated during the RareAlert development stage through supervised distillation, allowing the final model to internalize both integrated risk signals and diverse reasoning patterns.

Data partitioning and validation strategy

To prevent information leakage and to enable unbiased performance assessment, data splitting was predefined. Before model training, the dataset was split using stratified random sampling. Specifically, 80% of cases were randomly selected as the development set and the remaining 20% as the test set, separately within the rare disease and non – rare disease groups. All model selection and optimization procedures were confined to the development set.

Three-stage model development framework

A three-stage model development framework was adopted to integrate LLM-derived risk scores using machine learning.

In Stage 1 (broad screening), we aimed to identify promising combinations of LLM-derived features and learning algorithms. Feature sets ranging from single-model risk scores to multi-model aggregations were systematically evaluated. A diverse set of machine learning algorithms was considered, including logistic regression, random forest, extremely randomized trees, AdaBoost, gradient boosting methods (XGBoost and LightGBM), and CatBoost. Model performance was assessed using the area under the receiver operating characteristic curve (ROC – AUC), estimated via stratified cross-validation within the development set. Feature – algorithm combinations were ranked by AUC, and the top-performing candidates were selected for further evaluation.

In Stage 2 (model optimization and robustness assessment), the five highest-ranking feature – algorithm combinations from Stage 1 were further examined. For each selected combination, hyperparameters were optimized using an inner cross-validation loop, and model performance was assessed on outer validation folds to evaluate robustness and reduce overfitting.

In Stage 3 (final training and independent testing), the optimal model configuration identified in Stage 2 was trained on the full development dataset using the optimized hyperparameters. The finalized model was then evaluated once on the independent test set, which had not been used in any stage of model selection or hyperparameter tuning. The trained model took LLM-derived features as input and output predicted probabilities ranging from 0 to 1, with higher values indicating greater rare disease risk. These probabilities were multiplied by 100 to generate risk scores. Final model performance on the test set was quantified using ROC – AUC.

Importance analysis of different LLMs in the prediction task

Based on the final machine learning model, we performed an interpretability analysis to examine how different LLM-derived inputs contributed to rare disease risk prediction. The goal of this analysis was to obtain a transparent and quantitative understanding of how individual input features influence model outputs. Specifically, we aimed to identify which features were most influential overall, whether their effects tended to increase or decrease predicted risk, how dominant they were relative to other features, and how their contributions varied at both the population and individual levels.

To achieve these objectives, we applied SHAP (SHapley Additive exPlanations) analysis to decompose the model’s predicted risk probabilities into additive contributions from individual input features. This procedure yielded SHAP values for each feature and each sample.

Several complementary metrics were derived from the SHAP values. Normalized global importance was defined as the proportion of the total absolute SHAP attribution accounted for by each feature across all samples. This metric reflects the relative contribution of each feature to overall model predictions and allows feature importance to be compared on a standardized scale. Global SHAP value was calculated as the mean signed SHAP value for each feature across samples, reflecting the average directional effect of each LLM on the predicted risk. Positive values indicate a tendency to increase the predicted rare disease risk, whereas negative values indicate a tendency to decrease it. Finally, dominant-feature frequency was calculated by identifying, for each sample, the feature with the largest absolute SHAP value and computing the frequency with which each feature emerged as the primary driver of individual predictions.

**Stage 3: RareAlert training**

At this stage, we developed RareAlert as a single, deployable model designed to perform rare disease risk screening directly from free-text information. The RareAlert model was required to fulfill four key objectives: (i) to process unstructured clinical narratives and output quantitative rare disease risk scores with accompanying clinical explanations; (ii) to internalize heterogeneous medical reasoning patterns expressed by different LLMs; (iii) to learn the improved and calibrated risk estimates produced through machine learning–based integration of multi-LLM outputs; and (iv) to support efficient local deployment for large-scale screening while preserving data privacy.

RareAlert training can be viewed as a process of knowledge distillation and reasoning alignment, in which predictive signals and reasoning structures from a heterogeneous expert ensemble are compressed into a single deployable model. We adopted a supervised fine-tuning strategy based on an open-source large language model (Qwen3-4B). Crucially, the training targets for RareAlert were not derived from manual annotation or direct LLM judgment, but from the calibrated risk scores generated by the machine learning ensemble developed in the previous stage. This design reflects the practical difficulty of assigning accurate rare disease risk scores through expert labeling and leverages the superior performance of the ensemble model as a reliable supervisory signal.

Training data construction

Training data were constructed from cases in the development set. Each case was represented as a single training instance, with the input (user content) consisting of the original primary visit information, including demographic data, medical history, and physical examination findings. The output (assistant content) consisted of two components: (i) the medical reasoning paths generated by multiple LLMs for the given case, and (ii) the final reference risk score predicted by the machine learning ensemble model.

By incorporating ensemble-derived risk scores into the training targets, RareAlert was explicitly trained to align its predictions with the integrated multi-model consensus rather than with any individual LLM or subjective human judgment. This approach allowed the model to internalize a calibrated notion of rare disease risk that reflects complementary reasoning signals across models.

The contribution of individual LLMs to the ensemble prediction was quantified using SHAP analysis in the preceding machine learning stage. For each case, SHAP values were used to rank the relative importance of risk scores produced by different LLMs, with each LLM output treated as a distinct reasoning path. Based on this ranking, we systematically evaluated training configurations that incorporated between one and ten reasoning paths per case.

When a single reasoning path was included, the path corresponding to the LLM with the highest SHAP importance for that case was selected. For configurations incorporating n reasoning paths, the top n LLM outputs ranked by SHAP importance were included. This procedure enabled a data-driven, case-specific selection of reasoning supervision signals, ensuring that the most influential reasoning patterns identified by the ensemble model were preferentially learned during fine-tuning.

We fine-tuned Qwen3-4B and Qwen3-8 using supervised fine-tuning. All experiments were implemented with the LLaMAFactory framework and conducted on a total of 8 NVIDIA A800 80GB GPUs. We performed parameter-efficient fine-tuning with LoRA (rank r=8). The models were optimized using token-level cross-entropy loss with AdamW, a learning rate of $1 \times 10{-}5$, and a cosine learning-rate scheduler. Training was run for 3 epochs with a warm-up ratio of 0.1. The per-GPU batch size was set to 2, resulting in a total batch size of 16 across all GPUs.

**Stage 4: RareAlert validation**

In this stage, we evaluated whether the performance of RareAlert exceeded that of the machine learning models developed in Stage 2, individual LLMs, alternative ensemble techniques, and a multi-agent discussion framework. All evaluations were conducted on the independent test set, which comprised a stratified random sample of 20% of both rare disease and non-rare disease cases and was not used in any stage of machine learning model development or RareAlert training.

Model performance was assessed using the area under the receiver operating characteristic curve (ROC - AUC) as the primary metric. In addition, based on the optimal threshold determined using the Youden index, we evaluated sensitivity, specificity, false negative rate (FNR), false positive rate (FPR), positive predictive value (PPV), and negative predictive value (NPV).

Comparison with machine learning models from Stage 2

RareAlert was directly compared with the machine learning models developed during Stage 2 to assess whether supervised distillation into a single deployable model led to further improvements in predictive performance.

Comparison with individual LLMs

In addition to the LLMs used during system development, we included several additional large language models for comparative analysis, including Claude-3.7-Sonnet, DeepSeek-R1, Gemini-2.5-Pro, GPT-5, o3-mini, and Qwen3‑235B.

Comparison with alternative ensemble techniques

We further evaluated commonly used ensemble approaches by aggregating LLM-generated risk scores for each case using the mean, median, and mode. These aggregated scores were treated as final risk scores, and their predictive performance was assessed using the same evaluation metrics.

Comparison with a multi-agent discussion framework

We also assessed the performance of a multi-agent collaborative system consisting of three physician agents. The system received primary-visit clinical information as input, and the three agents engaged in structured discussion and debate to reach a consensus risk score. Because multi-agent discussions involve substantial token consumption and high computational and API costs, this analysis was conducted on a randomly selected subset of 1,000 cases.

**External validation**

RareAlert and all comparative AI systems were further evaluated using an external validation dataset. All systems included in the comparative analyses were tested on this external dataset, which underwent the same data screening, deduplication, and structuring procedures as the internal dataset.

**Reliability Analysis**

To assess the reliability of LLM predictions for rare disease, we conducted a reliability analysis. Given the large volume of data in this study, the reliability analysis utilized the 750 cases from the pilot study. All included models were tested 10 times on these cases. We recorded the scores from each iteration and analyzed inter-run agreement using Fleiss' kappa coefficient. The results are listed in supplementary file.